\pdfoutput=1
\documentclass[11pt]{article}

\usepackage{EACL2023}

\usepackage{times}
\usepackage{latexsym}
\usepackage{graphicx}
\usepackage[T1]{fontenc}
\usepackage{hyperref}
\usepackage[utf8]{inputenc}

\usepackage{microtype}
\usepackage{tabularx}
\usepackage{booktabs}
\usepackage{inconsolata}
\usepackage{todonotes}
\newcommand{\ggrev}[2]{\textcolor{red}{#2}}

\title{UNSEE: Unsupervised Non-contrastive Sentence Embeddings}

\author{Ömer Veysel Çağatan \\
    Koç University \\
   Rumelifeneri, Sarıyer Rumeli Feneri Yolu \\
   34450 Sarıyer/İstanbul,Turkey\\
  \texttt{ocagatan19@ku.edu.tr}}

\begin{document}
\maketitle
\begin{abstract}
We present UNSEE: Unsupervised Non-Contrastive Sentence Embeddings, a novel approach that outperforms SimCSE in the Massive Text Embedding benchmark. Our exploration begins by addressing the challenge of representation collapse, a phenomenon observed when contrastive objectives in SimCSE are replaced with non-contrastive objectives. To counter this issue, we propose a straightforward solution known as the target network, effectively mitigating representation collapse. The introduction of the target network allows us to leverage non-contrastive objectives, maintaining training stability while achieving performance improvements comparable to contrastive objectives. Our method has achieved peak performance in non-contrastive sentence embeddings through meticulous fine-tuning and optimization. This comprehensive effort has yielded superior sentence representation models, showcasing the effectiveness of our approach.
\end{abstract}

\section{Introduction}

\begin{figure*}[htbp]
  \centering
  \includegraphics[width=1\textwidth]{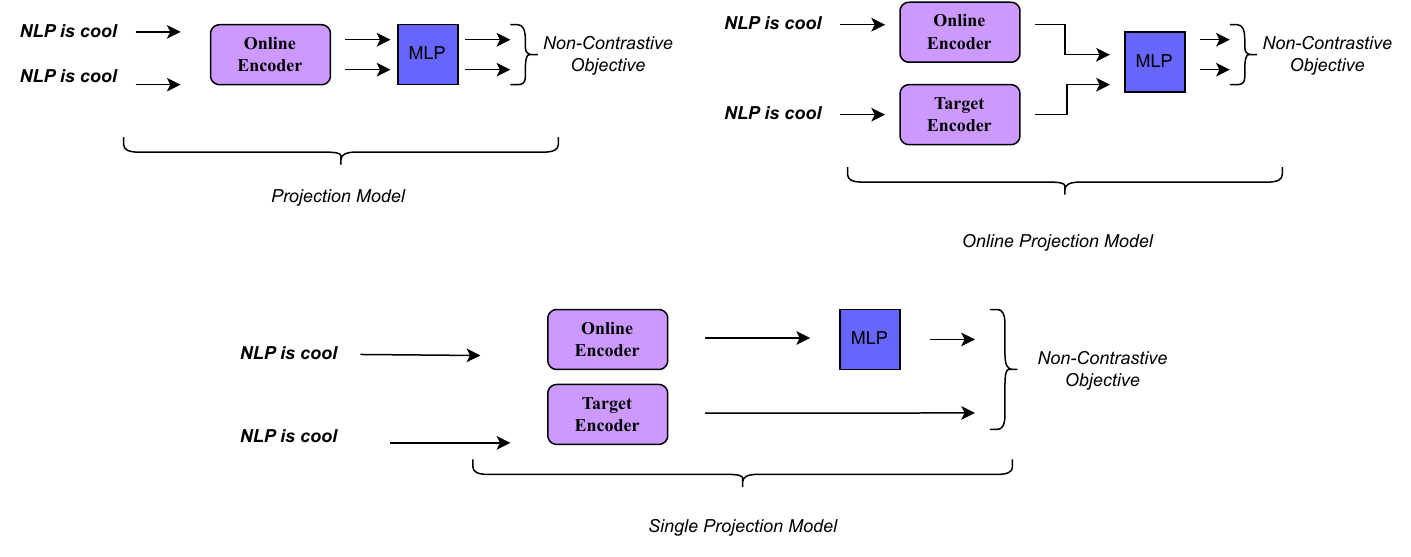} 
  \caption{ {\sl Projection Model} is the same as SimCSE~\cite{Gao2021SimCSESC}. The Online keyword is to emphasize that the model gets gradient updates. The {\sl Online Projection Model} is similar to the {\sl Projection Model} except for the Target Encoder. The Target Encoder is an exponentially moving average of the Online network. Both outputs from Online and Target Encoders pass through the same MLP layer in the Online Projection Model. Target MLP is not employed due to the nature of fine-tuning which will slightly change the newly initialized MLP layer that will potentially corrupt the embeddings. In {\sl Single Projection Model}, Target embeddings do not go through the MLP layer unlike {\sl Online Projection Model}. {\sl Single Projection Model} is identical to the architecture proposed in BSL~\cite{Zhang2021BootstrappedUS}. We only use BERT-base ~\cite{devlin2018bert} as the encoder.}
  \label{fig:arc}
\end{figure*}

Contrastive learning has been used quite extensively in the sentence embedding models~\cite{zhang-etal-2021-bootstrapped,liu-etal-2021-fast,reimers-2019-sentence-bert,chuang-etal-2022-diffcse,Gao2021SimCSESC,jiang2022promcse, Liu2022} which hace achieved remarkable results on MTEB benchmark~\cite{muennighoff-etal-2023-mteb}. The fundamental role of the contrastive objective is to regularize the anisotropic embedding space of language models, ultimately enabling them to function effectively as embedding models~\cite{li-etal-2020-sentence}.

On the contrary, non-contrastive methods have not gained widespread popularity as the primary objective for training sentence embedding models, despite demonstrating regularization efficacy in vision~\cite{bardes2022vicreg,Zbontar2021BarlowTS,Chen2020ExploringSS,Grill2020BootstrapYO}. This reluctance stems from the fact that non-contrastive objectives tend to perform suboptimally in comparison to contrastive objectives, particularly in the SimCSE~\cite{Gao2021SimCSESC} setting. For example, SCD~\cite{Klein2022SCDSD} showcased that Barlow Twins~\cite{Zbontar2021BarlowTS} achieves only 67.57 on the STSBenchmark~\cite{cer-etal-2017-semeval} test set, while SimCSE~\cite{Gao2021SimCSESC} accomplishes 76.85.

\begin{figure*}[htbp]
  \centering
  \includegraphics[width=1\textwidth]{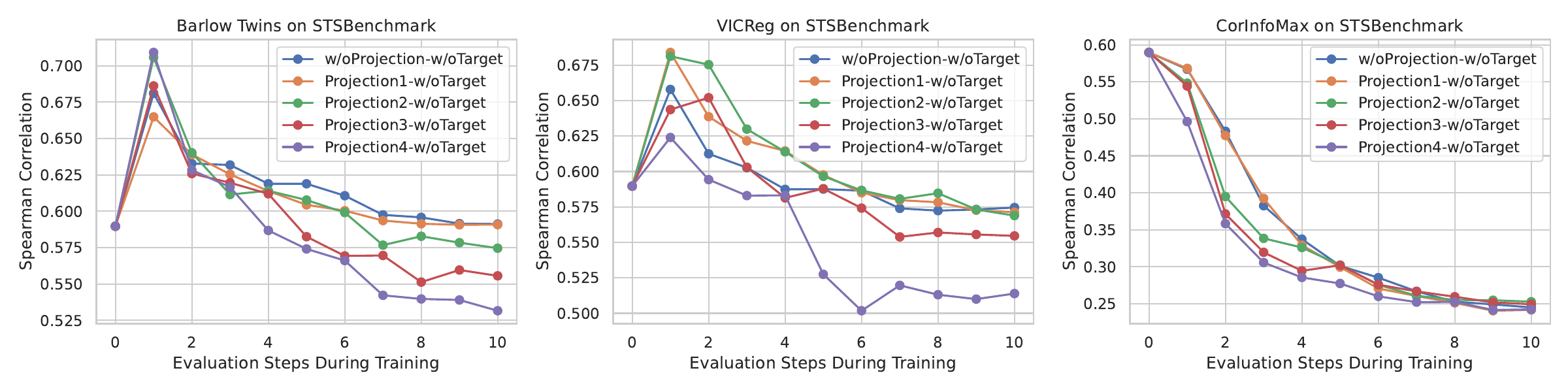} 
  \caption{The performance of various non-contrastive objectives on STSBenchmark evaluation dataset~\cite{cer-etal-2017-semeval} in the Projection Model or SimCSE setting. The difference between models is the number of MLP layers. MLP layer is adopted from BSL~\cite{zhang-etal-2021-bootstrapped}.}
  \label{fig:collapse}
\end{figure*}

Additionally, we demonstrate that the observed performance drawback is not confined to Barlow Twins exclusively. Other well-known non-contrastive methods~\cite{bardes2022vicreg,ozsoy2022selfsupervised} also suffer from inferior performance. Specifically, when examining the top evaluation scores in Figure \ref{fig:collapse} for the STSBenchmark development set, these non-contrastive methods consistently fall short compared to SimCSE, which achieves an impressive score of 82.5.

Despite the comparatively lower performance observed when non-contrastive objectives are employed in a sentence embedding framework, their inherent characteristics, such as the lack of negative samples and the ability to prevent dimensional collapse, as demonstrated in~\citet{ozsoy2022selfsupervised}, inspire us to delve deeper into investigating and improving the effectiveness of non-contrastive objectives.

Hence, we begin by presenting empirical evidence of representation collapse observed during training with non-contrastive objectives. This includes instances utilizing siamese networks, dropout as augmentation, and even those incorporating additional parametrization with MLP layers. We delve into the potential reasons behind the suboptimal performance in Section \ref{sec:projmodel}.

Furthermore, we introduce the target network as a novel augmentation method, which empirically enhances the diversity of embeddings and effectively mitigates the collapse associated with non-contrastive objectives. Subsequently, through additional finetuning and architectural refinements, detailed in Section \ref{sec:onlineprojmodel} and Section \ref{sec:singleprojmodel}, we achieve the absolute best performance among non-contrastive objectives. In summary, we present a series of non-contrastive models collectively named UNSEE, surpassing SimCSE in the MTEB benchmark. This underscores the potential of non-contrastive objectives as fundamental components for training state-of-the-art embedding models.

\section{Related Work}

Competitive sentence embedding models are typically built by modifying BERT~\cite{devlin2018bert} with diverse configurations. In the early stages of sentence embedding development, models like InferSent~\cite{conneau-etal-2017-supervised} and the Universal Sentence Encoder~\cite{cer-etal-2018-universal} predominantly relied on LSTM~\cite{Hochreiter1997LongSM} or the Transformer~\cite{Vaswani2017AttentionIA} architecture.

The conventional BERT model~\cite{devlin2018bert} exhibits suboptimal performance and operates at a slower pace. Sentence BERT, abbreviated as SBERT~\cite{reimers-2019-sentence-bert}, represents a modified version of BERT that utilizes siamese or triplet networks to generate meaningful and accurate sentence embeddings. SBERT improves accuracy and significantly reduces the time required to identify the most similar pair of sentences within a set of 10,000 sentences, reducing the process from 65 hours to just 5 seconds. Despite the integration of these enhancements into BERT, a fundamental question arises: why are these modifications necessary in the first place?

\citet{li-etal-2020-sentence} brings attention to a concern related to BERT's sentence embeddings, specifically highlighting the presence of anisotropy in the embedding space. Their empirical observations reveal that the sentence embedding space lacks smoothness and is poorly defined in certain regions, posing challenges when applying cosine similarity directly. To address this issue, they propose a solution that involves transforming sentence embeddings into a Gaussian distribution that is both smooth and isotropic. This transformation is achieved through the utilization of normalizing flows. The proposed flow-based generative model is trained in an unsupervised manner with the objective of maximizing the likelihood of generating BERT sentence embeddings from a standard Gaussian latent variable.

\citet{liu-etal-2021-fast} present MirrorBERT, a method that improves sentence representations through a straightforward approach of duplicating or slightly augmenting the text input, all without external supervision. These augmentations can take place either within the input space, involving actions like random span masking, or within the feature space, using techniques such as dropout. Notably, dropout is not only implemented within the MLP but also leads to the deactivation of attention heads, all while preserving the model's performance across various tasks. Furthermore, it has been demonstrated that MirrorBERT also enhances isotropy.

\citet{Gao2021SimCSESC} introduce SimCSE, which employs conventional dropout as a means of input augmentation. By feeding a single sentence through two passes, this approach generates two distinct feature embeddings, which can be treated as similar to positive pairs, while other sentences serve as negative samples. This dropout-based approach offers a straightforward technique for creating positive-negative pairs in contrastive learning. Impressively, it achieves superior performance compared to Mirror-BERT with only moderate modifications.

The current state-of-the-art embedding models~\cite{bge_embedding,li2023general,su2023embedder,Wang2022TextEB} distinguish themselves by their training on exceptionally large and extensive corpora. These corpora encompass a vast amount of both unlabeled and labeled text data. The utilization of such extensive and diverse training data has played a crucial role in the impressive performance exhibited by these models in the MTEB benchmark~\cite{muennighoff-etal-2023-mteb}, despite their fundamental similarity to SimCSE.

On the contrary, models such as SimCSE follow a significantly different paradigm, undergoing training on a relatively modest dataset consisting of just 1 million sentences. Considering the substantial difference in the scale and diversity of training data, attempting direct comparisons between SimCSE-like models and these state-of-the-art embedding models seems impractical and might not provide meaningful insights into their relative capabilities. Therefore, we exclude them from our analysis.

\begin{figure*}[htbp]
  \centering
  \includegraphics[width=1\textwidth]{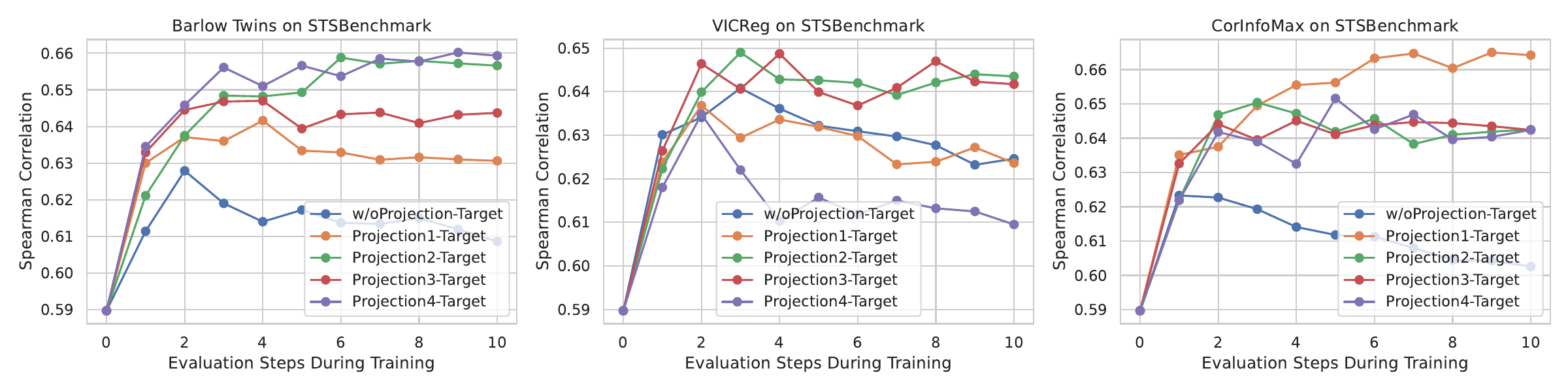} 
  \caption{The performance of various non-contrastive objectives on STSBenchmark~\cite{cer-etal-2017-semeval} in the Online Projection Model with SimCSE hyperparameters. The difference between models is the number of MLP layers. MLP layer is adopted from BSL~\cite{zhang-etal-2021-bootstrapped}.}
  \label{fig:target}
\end{figure*}
    
\section{Background}
In this section, we provide an extensive overview of non-contrastive representation learning and the methods that form the core of our research.

\subsection{Non-Contrastive Representation Learning}

Recent advancements in the field of self-supervised visual learning have extended beyond the traditional contrastive approach, exploring innovative avenues that reduce the reliance on negative sample pairs. These methods primarily focus on enhancing the quality of independently augmented representations, forming a subset of non-contrastive frameworks. To address challenges such as model collapse, various effective strategies have emerged within this domain. These include the adoption of asymmetric network architectures~\cite{Grill2020BootstrapYO, Chen2020ExploringSS}, feature decorrelation techniques~\cite{Zbontar2021BarlowTS,bardes2022vicreg,ozsoy2022selfsupervised, Ermolov2020WhiteningFS}, as well as clustering methods~\cite{Amrani2021SelfSupervisedCN,Assran2022MaskedSN,caron2019deep,Caron2020UnsupervisedLO}, all of which contribute to the progress in self-supervised visual learning while addressing the challenges inherent to this domain.

\subsection{CorInfoMax}
CorInfoMax~\cite{ozsoy2022selfsupervised} utilizes a second-order statistics-based mutual information measure to gauge the level of correlation among its input components. The primary aims of maximizing this measure between different representations of the same input are twofold: firstly, it mitigates the risk of feature vector collapse by generating feature vectors with non-degenerate covariances. Secondly, it establishes relevance among these alternative representations by enhancing their linear interdependence. 

An approximation of this information maximization objective simplifies into an Euclidean distance-based objective function, which is further regulated by the logarithm of the determinant of the feature covariance matrix. This regularization term serves as a natural safeguard against feature space degeneracy. Consequently, the proposed approach not only prevents complete output collapse to a single point but also effectively averts dimensional collapse by encouraging the dispersion of information across the entire feature space.

\subsection{Barlow Twins}
The Barlow Twins~\cite{Zbontar2021BarlowTS} is designed to prevent collapse naturally. It accomplishes this by assessing the cross-correlation matrix between the outputs of two identical networks, which are fed with altered versions of a sample. The goal is to make this cross-correlation matrix as similar to the identity matrix as possible. Consequently, this approach ensures that the embedding vectors of these distorted sample versions become more alike, all while reducing redundancy among their components. Importantly, Barlow Twins operates without the need for large batch sizes or introducing any disparities between the network twins, such as the inclusion of a predictor network, gradient stopping, or utilizing a moving average for weight updates.

\begin{figure*}[htbp]
\centering
\includegraphics[width=1\textwidth]{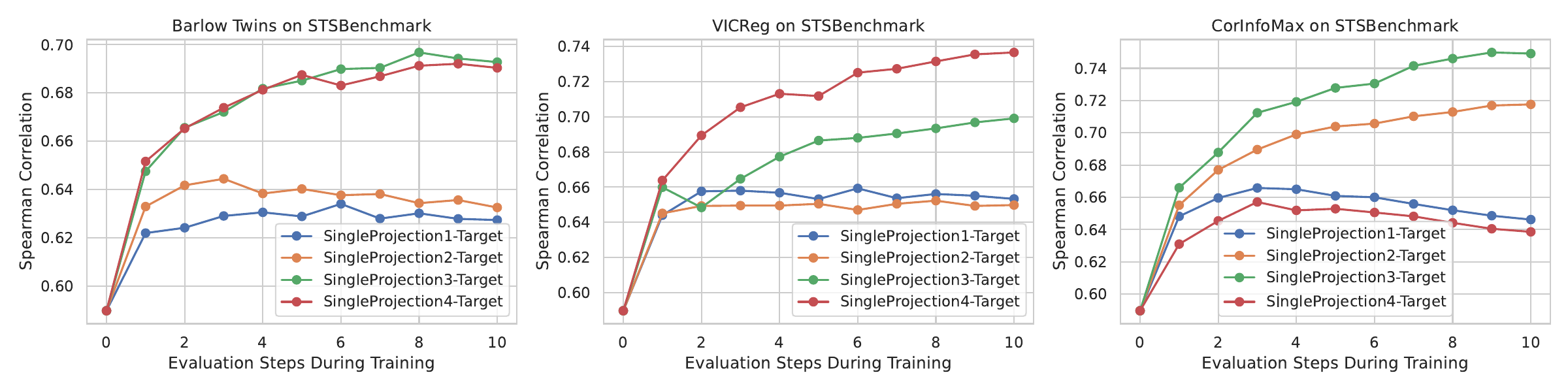}  
\caption{The performance of various non-contrastive objectives on STSBenchmark~\cite{cer-etal-2017-semeval} in the Single Projection Model with SimCSE hyperparameters. \ggrev{The difference between models is the number of MLP layers. MLP layer is adopted from BSL~\cite{zhang-etal-2021-bootstrapped}.}{}}
\label{fig:singlepro}
\end{figure*}

\subsection{VICReg}
VICReg~\cite{bardes2022vicreg}, short for Variance-Invariance-Covariance Regularization, is an approach specifically designed to address the issue of collapse straightforwardly. It accomplishes this by introducing a simple regularization term that focuses on the variance of the embeddings along each dimension individually. In addition to the variance component, VICReg incorporates a mechanism that reduces redundancy and ensures decorrelation among the embeddings, achieved through covariance regularization.

\subsection{BYOL}
BYOL~\cite{Grill2020BootstrapYO} hinges on the utilization of two distinct neural networks, namely the online and target networks, which collaborate and mutually enhance their learning processes. This technique operates by presenting an augmented view of an image to the online network, to train it to predict the representation of the same image as processed by the target network but under a different augmented view. Simultaneously, the target network undergoes updates through a slow-moving average mechanism based on the evolving state of the online network.

This approach essentially fosters a dynamic interplay between the online and target networks, where they iteratively adapt and refine their representations in response to the variations in augmented views. Through this collaborative learning process, BYOL aims to yield highly informative and generalized feature representations, making it particularly valuable for self-supervised learning tasks, where labeled data may be limited or unavailable.

\section{From SimCSE to the UNSEE}

In this section, we detail the methodology employed to derive the final UNSEE models from SimCSE. The STSBenchmark evaluation dataset~\cite{cer-etal-2017-semeval} serves as the basis for identifying the optimal configuration. We follow a systematic approach, progressively discussing enhancements and offering justifications for each decision. It's worth noting that SimCSE achieves a score of 82.5 in the STSBenchmark. However, we intentionally exclude it from our figures as its high score can distort the visualization in certain experiments.

\subsection{Projection Model}\label{sec:projmodel}

In Figure~\ref{fig:arc}, \textit{Projection Model} corresponds to the precise configuration outlined in SimCSE~\cite{Gao2021SimCSESC}, wherein dropout serves as a straightforward augmentation technique.

Figure \ref{fig:collapse} offers compelling evidence of substantial deficiencies in non-contrastive models when employed within the SimCSE framework. It's conceivable to assert that these models undergo a representation collapse during their training phase. This leads to critical questions regarding the broader versatility and generalization capacity of such objectives, hinting at their potential effectiveness within constrained domains or contexts.

Conversely, it is noteworthy that dropout augmentation plays a pivotal role within the SimCSE paradigm. This realization leads us to consider the prospect of exploring alternative augmentation techniques, aiming to delve deeper into the inherent potential of non-contrastive objectives. This exploration of diverse augmentation strategies has the potential to reveal the true efficacy and versatility of these objectives, providing insights into their capabilities beyond their current limitations.
\begin{figure*}[htbp]
  \centering
  \includegraphics[width=1\textwidth]{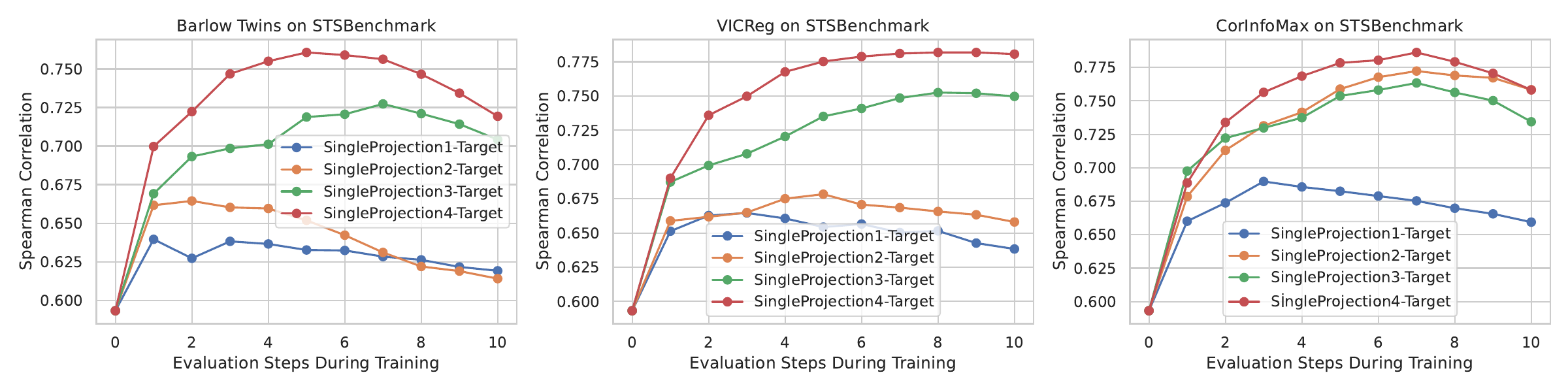}  
  \caption{The performance of various non-contrastive objectives on STSBenchmark~\cite{cer-etal-2017-semeval} in the Single Projection Model with slightly optimized hyperparameters. The difference between models is the number of MLP layers. MLP layer is adopted from BSL~\cite{zhang-etal-2021-bootstrapped}.}
  \label{fig:singleprobsl}
\end{figure*}

\subsection{Online Projection Model} \label{sec:onlineprojmodel}

Considering the notable underperformance of non-contrastive objectives, it becomes imperative to explore novel avenues for their improvement. As highlighted by ~\citet{gao-etal-2021-making}, most input space augmentations are not as effective as dropout. This finding casts doubt on the likelihood of discovering an input augmentation method superior to dropout.

This recognition has guided our exploration towards the creation of a new augmentation technique, specifically, the incorporation of a target network. This method constitutes a relatively straightforward feature space augmentation strategy aimed at infusing greater diversity into the embeddings, surpassing the effectiveness of conventional dropout. An analogy can be drawn to \textit{lagged dropout}, where networks undergoing dropout display subtle variations, and the target network functions as a slow-moving average of the online network, actively contributing to the diversification of embeddings.

Figure \ref{fig:target} demonstrates that the utilization of a target network effectively prevents representation collapse, ensuring a more stable training process. However, it is noteworthy that, even in situations where representation collapse is avoided, the overall performance remains suboptimal. The introduction of additional parametrization through MLP layers has only yielded a marginal impact on improving performance.

An argument can be made that creating effective sentence embeddings presents a more formidable challenge when non-contrastive objectives are utilized, especially in comparison to tasks related to vision. In contrastive learning, the approach involves actively pushing data samples apart to improve discrimination. However, in sentence embeddings with non-contrastive objectives, this process becomes implicit.

To draw a parallel, envision a scenario where each sample is assigned a distinct label, yet some labels are shared among the samples. Similarly, when training a sentence embedding model with non-contrastive objectives, it reflects this intricate situation. We utilize a dataset consisting of randomly sampled Wikipedia sentences collected in SimCSE~\cite{Gao2021SimCSESC}. While each sentence in the dataset may possess unique content, there exist underlying semantic or syntactic relationships among them, akin to the shared labels in the problem we are considering. The inherent complexity and the necessity to implicitly capture these relationships contribute to the intricacy of the sentence embedding task when utilizing non-contrastive objectives.

\begin{table*}[t!]
    \centering
    \resizebox{0.9\textwidth}{!}{\begin{tabular}{l|ccccccccc}
    \toprule
 & Class. & Clust. & PairClass. & Rerank. & Retr. & STS & Summ. & Avg. \\
Num. Datasets ($\rightarrow$) & 12 & 11 & 3 & 4 & 15 & 10 & 1 & 56 \\
\midrule
\midrule
\multicolumn{9}{l}{\emph{Self-supervised methods}}
\\
\midrule
Glove & 57.29 & 27.73 & 70.92 & 43.29 & 21.62 & 61.85 & 28.87 & 41.97 \\
Komninos & 57.65 & 26.57 & \textbf{72.94} & 44.75 & 21.22 & 62.47 & 30.49 & 42.06 \\
BERT & 61.66 & 30.12 & 56.33 & 43.44 & 10.59 & 54.36 & 29.82 & 38.33 \\
SimCSE & 62.50 & 29.04 & 70.33 & 46.47 & 20.29 & \textbf{74.33} & \textbf{31.15} & 45.45 \\
UNSEE-BYOL(Ours) & 62.55 & 27.81 & 65.3 & 46.47 & 23.11 & 73.04 & 30.68 & 45.46 \\
UNSEE-Barlow(Ours) & 62.76 & \textbf{30.04} & 65.7 & 46.9 & 23.06 & 72.15 & 30.25 & 45.82 \\
UNSEE-CorInfoMax(Ours) & \textbf{62.85} & 28.90 & 67.87 & 46.81 & \textbf{24.80} & 72.31 & 30.81 & 46.22 \\
UNSEE-VICReg(Ours) & 62.58 & 28.44 & 70.24 & \textbf{47.23} & 24.79 & 73.11 & 30.34 & \textbf{46.37} \\
    \bottomrule
    \end{tabular}}
    \caption{Average of the main metric from ~\citet{muennighoff-etal-2023-mteb} per task per model on MTEB English subsets. SimCSE, BERT, Komnimos, and Glove scores are taken from ~\citet{muennighoff-etal-2023-mteb}}
    \label{tab:results}
\end{table*}

\subsection{Single Projection Model} \label{sec:singleprojmodel}

In our \textit{Online Projection Model}, it is crucial to emphasize the significant contribution of MLP layers for both target and online embeddings. Importantly, the sentence embeddings themselves are initially obtained from the BERT model.

The MLP layers should not be viewed as static components in our model architecture; instead, they play a dynamic and transient role during the training phase. Their function is crucial in continually shaping the embeddings for effective loss minimization. However, it is important to emphasize that the outputs produced by these MLP layers do not represent the definitive embeddings used for subsequent evaluation.

This leads us to an intriguing hypothesis: What if we were to consider avoiding the involvement of MLP layers in the processing of the target network's embeddings? By establishing a direct, unmediated connection between the loss minimization process and the generation of embeddings, we aim to explore whether such architectural simplification could yield substantial advantages. This modification holds the potential to provide insights into whether a more simplified approach might enhance both the efficiency of loss minimization and the quality of the resultant embeddings, thereby refining the overall training process.

The outcomes presented in Figure \ref{fig:singlepro} closely align with our hypothesis. Throughout the training process, the models consistently showcased incremental performance improvements, surpassing the accomplishments of the preceding model while maintaining identical complexities and hyperparameters. While these results are undeniably promising, it is crucial to acknowledge that they have not yet reached the performance level observed in SimCSE. This suggests that additional optimization endeavors are necessary to narrow the gap and enable our models to attain the performance parity with their SimCSE counterparts. Hence, there is ample room for refinement and enhancement in our pursuit of achieving comparable or even superior performance.

We have significantly improved our model's performance by making relatively minor adjustments to specific hyperparameters, with a particular focus on the learning rate, batch size, and sequence length. The optimal hyperparameters are set to 1e-4, 32, and 64, respectively. The decay rate is maintained at 0.999 consistently across all experiments. Remarkably, these subtle modifications have enabled us to achieve the highest attainable scores among non-contrastive objectives, all without delving into the optimization of hyperparameters within the loss objective. It's important to note that we intentionally adhered to default values for the objectives, highlighting the robustness and transferability of these objectives across different domains. This observation underscores the versatility of the objectives, demonstrating their effective performance even when applied in contexts beyond their original domain.

The outcomes shown in Figure \ref{fig:singleprobsl} do not signify the peak of our accomplishments. We have obtained superior results by increasing the frequency of evaluations (20 evaluations per run) throughout the training process and introducing a checkpointing system to preserve the best-performing model. These particular runs were crafted to be consistent with our earlier experiments, intending to showcase the effectiveness of the implemented adjustments.

\section{Evaluation Dataset}
\subsection{MTEB Benchmark}

The primary goal of the Massive Text Embedding Benchmark (MTEB)~\cite{muennighoff-etal-2023-mteb} is to offer a comprehensive assessment of model performance across a diverse range of text embedding tasks. It serves as a valuable resource for identifying text embeddings that exhibit universal applicability across a wide spectrum of tasks. MTEB encompasses an extensive collection of 58 datasets spanning 112 languages, encompassing 8 distinct embedding tasks, including bitext mining, classification, clustering, pair classification, reranking, retrieval, STS (Semantic Textual Similarity), and summarization.

\section{BYOL, BSL and Final Results}
In our paper, we extensively examine and engage in discussions concerning non-contrastive objectives that incorporate a siamese network architecture. However, it's important to note that our most effective configuration closely resembles BYOL~\cite{Grill2020BootstrapYO}, and we have conducted training to incorporate this configuration into our results. The ultimate model we present is a variation of BSL~\cite{zhang-etal-2021-bootstrapped} with dropout serving as an augmentation method.

Throughout our experimentation, it becomes evident that non-contrastive methods consistently outperform SimCSE as the table \ref{tab:results} verifies. The degree of improvement varies, with some methods showing only marginal enhancements, while others exhibit significantly more substantial gains. This overarching pattern underscores the compelling impact of non-contrastive objectives on augmenting BERT's proficiency as a sentence embedding model.

While MTEB aims to encompass a wide range of applications for sentence embeddings, there are noticeable score discrepancies within UNSEE models. Despite their shared objective of optimizing feature decorrelation, implicit in the case of BYOL, differences in their problem formulations lead to variations in scores across different subtasks. For instance, UNSEE-Barlow excels significantly in clustering compared to other objectives. One could argue that the exclusive focus of Barlow Twins on minimizing feature decorrelation might make it more effective in information dissemination, resulting in superior clustering. However, VICReg's incorporation of variance and invariance aspects may pose challenges in achieving the same level of clustering performance. Another question arises regarding why this performance difference doesn't extend to retrieval. One possible explanation is that retrieval requires a finer-grained spread within a subspace, a quality that other objectives (excluding Barlow Twins) may achieve due to their invariance objective.

Nonetheless, our findings collectively reinforce the notion that non-contrastive methods contribute to a notable expansion of BERT's capabilities, effectively harnessing its potential to serve as a highly effective and versatile tool for generating sentence embeddings. This empirical evidence underscores the transformative role these methods play in enhancing the utility and adaptability of BERT across various sentence-related tasks.

\section{Conclusion}

 UNSEE (Unsupervised Non-Contrastive Sentence Embeddings) is a simple framework for non-contrastive sentence embeddings, which outperforms SimCSE in the Massive Text Embedding Benchmark (MTEB). We address representation collapse using a simple solution called the target network, enabling stable training and achieving performance similar to contrastive objectives. Our meticulous fine-tuning leads to performant sentence embedding models, showcasing the significance of thoughtful optimization in advancing non-contrastive methods for sentence representation.

\section*{Limitations}
UNSEE models have inherent limitations stemming from their training data, which encompasses only one million sentences. In contrast, state-of-the-art embedding models undergo training on datasets comprising over a hundred million, or even more than a billion pairs. As a result, our models are expected to exhibit inferior performance when compared to models specifically designed for sentence embedding. We recommend considering the top-performing models on the MTEB leaderboard for more effective practical use.

\section*{Ethics Statement}
The models under examination, UNSEE-*, lack generative abilities, ensuring their incapacity to produce unfair, biased, or harmful content. The datasets utilized in this study have been meticulously selected from reputable repositories known for their safety in research applications, with strict measures in place to prevent the inclusion of personal information or offensive material.

\section*{Training Details}
We implement UNSEE with {\sl SentenceTransformers} from ~\cite{reimers-2019-sentence-bert}. Our \href{https://github.com/asparius/UNSEE}{code} is available at GitHub. To compare our models while developing them we keep the hyperparameters as same as the SimCSE which are 64 for batch size, 3e-5 for learning rate and 32 for the sequence length. When the target network is employed, the decay rate is 0.999 throughout all experiments. Our best models have 32 for the batch size, 1e-4 for the learning rate, and 64 for the sequence length, decay rate is the same. Best BYOL and VICReg models use 3 layers of MLP. CorInfoMax and Barlow Twins use 4. We use the same MLP architecture as BSL~\cite{zhang-etal-2021-bootstrapped}. In Barlow Twins, we use the same $\lambda$ as the original paper which is 0.0051. In VICReg, we use the same hyperparameter weights from the original paper which are 25 for invariance and variance, 1 for covariance. In CorInfoMax, we use R\_ini=1, la\_=0.01,la\_mu=0.01, R\_eps\_weight=1e-6, 0.2 for covariance and 2000 for invariance loss.

\section*{Computational Requirements}
We only use Tesla T4 GPUs for our experiments.

\section*{Acknowledgements}
We are grateful to Alper Erdogan, and Deniz Yuret for advising the project initially and hereby thank KUIS AI for providing computing resources for our project. 
\nocite{*}
\bibliography{eacl2023}
\bibliographystyle{eacl2023}

\end{document}